\documentclass[conference]{IEEEtran}
\usepackage{cite}
\usepackage{amsmath,amssymb}
\usepackage{graphicx}
\usepackage{url}
\usepackage{booktabs}
\usepackage{listings}
\usepackage{algorithm}
\usepackage{algpseudocode}

\title{ML-Based Automata Simplification \\for Symbolic Accelerators}

\author{
\IEEEauthorblockN{Tiffany Yu, Rye Stahle-Smith, Darssan Eswaramoorthi, Rasha Karakchi}
\IEEEauthorblockA{University of South Carolina\\
Email: \{tyu, ryes, darssan\}@email.sc.edu, karakchi@cec.sc.edu}
}

\begin{document}
\maketitle

\begin{abstract}
Symbolic accelerators are increasingly used for symbolic data processing in domains such as genomics, NLP, and cybersecurity. However, these accelerators face scalability issues due to excessive memory use and routing complexity, especially when targeting a large set. We present \textbf{AutoSlim}, a ML-based learning-based graph simplification framework designed to reduce the complexity of symbolic accelerators designed based on Non-deterministic Finite Automata (NFA) deployed on FPGA-based overlays such as NAPOLY+. AutoSlim uses Random Forest classification to prune low-impact transitions based on edge scores and structural features, significantly reducing automata graph density while preserving semantic correctness. Unlike prior tools, AutoSlim targets automated score-aware with weighted transitions, enabling efficient ranking-based sequence analysis. We evaluated data sets (1K–64K nodes) in NAPOLY+ and conducted performance measurements including latency, throughput, and resource usage. AutoSlim achieves up to 40\% reduction in FPGA LUTs and over 30\% pruning in transitions, while scaling to graphs an order of magnitude larger than existing benchmarks. Our results also demonstrate how hardware interconnection (fanout) heavily influences hardware cost and that AutoSlim’s pruning mitigates resource blowup. 
\end{abstract}


\section{Introduction}
Symbolic data, such as genomic sequences, behavioral logs, and tokens, appear widely in fields such as bioinformatics, cybersecurity, and NLP. These workloads often rely on finite automata, where patterns are modeled as graphs of states and transitions~\cite{woods2018automata, karakchi2025ai}. Automata offer a powerful abstraction for high-throughput pattern matching and symbolic analysis.

\begin{figure}[h]
\centering
\includegraphics[width=0.3\textwidth]{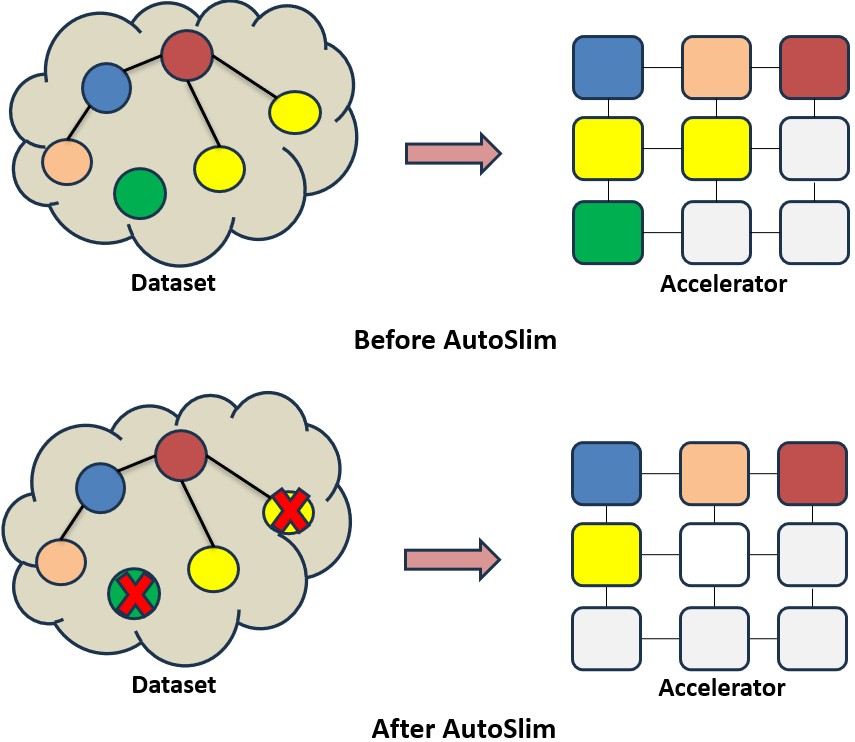}
\caption{Illustration of AutoSlim's graph optimization. \textit{Top:} Redundant nodes (yellow) and unreachable states (green) increase PE and routing cost. \textit{Bottom:} Pruned graph after AutoSlim removes redundancies.}
\label{fig:system}
\end{figure}

However, CPU and GPU automata processing suffers from limited parallelism and poor memory locality. FPGA-based accelerators provide higher throughput by parallelizing state transitions~\cite{karakchi2019overlay, karakchi2016high, karakchi2017reconfigurable, karakchi2024transformer, karbowniczak2024scored, karbowniczak2025optimizing}. Among these, \textbf{NAPOLY+} is a scalable NFA overlay that supports cost-aware matching by encoding transition weights and selecting minimal-cost paths~\cite{karakchi2023napoly, karbowniczak2024scored}. It maps ANML-described automata~\cite{wadden2016anmlzoo, Dlugosch14} directly to hardware using BRAMs for transitions and point-to-point connected state transition elements (STE). As symbolic data sets scale, large automata graphs pose several challenges:
(1) excessive memory usage from dense transitions,
(2) routing congestion due to rich interconnects, and
(3) redundant computation from unreachable or duplicate nodes.
To address these, we propose \textbf{AutoSlim}, a machine learning-based preprocessing tool that prunes automata graphs before hardware synthesis. It reduces graph size while preserving correctness, improving resource usage and execution efficiency on FPGAs. Figure~\ref{fig:system} illustrates the motivation of AutoSlim. The original automaton (top) contains duplicate destinations (yellow nodes) with identical scores and unreachable states (green). Mapping such graphs to hardware incurs an unnecessary logic cost. AutoSlim merges equivalent structures and removes inactive nodes, reducing resource usage without affecting output.

Real-world automata graphs exhibit significant redundancy—unused transitions, duplicated paths, and statistically insignificant states. Prior techniques like partitioning~\cite{nourian2017demystifying}, state merging~\cite{wadden2017automata}, or structural heuristics~\cite{karakchi2023napoly, karakchi2019overlay} often trade off interpretability or overlook probabilistic edge importance. AutoSlim uses a Random Forest classifier~\cite{breiman2001random} trained on annotated automata graphs to prune low-impact transitions based on edge frequency, cost, and node centrality. It preserves the semantics of scoring-based NFA execution in NAPOLY+, ensuring correctness in ranked pattern matching while improving hardware efficiency. Unlike static pruning, AutoSlim’s learning-driven strategy generalizes across workloads and application domains.

\section{Background and Related Work}

Finite automata are central to symbolic tasks such as pattern matching, intrusion detection, and sequence alignment~\cite{karakchi2023napoly, karakchi2024transformer}. While CPUs and GPUs suffer from irregular memory access and limited parallelism, spatial architectures, especially FPGAs and Micron's Automata Processor (AP), offer scalable alternatives by exploiting concurrent state transitions.

NAPOLY+ is an FPGA overlay for nondeterministic finite automata(NFA), representing automata as weighted graphs~\cite{karbowniczak2024scored, karbowniczak2025optimizing}. Each edge carries a symbolic transition and domain-specific cost, allowing ranked pattern matching through minimum-cost path selection. Unlike binary-matching automata processors, NAPOLY+ models automata as probabilistic graphs with flexible BRAM-based transitions and configurable State Transition Elements (STEs). While effective, large graphs introduce hardware bottlenecks including routing congestion and resource overhead.

Previous research has primarily focused on optimizing execution backends without modifying the automaton. Nourian \textit{et al.}~\cite{nourian2017demystifying} benchmark automata execution on GPUs, FPGAs, and Micron's AP, using unit-cost graphs from ANMLZoo~\cite{wadden2016anmlzoo}, but do not explore automaton simplification. Wadden \textit{et al.}~\cite{wadden2017automata} propose Automata-to-Routing to enhance spatial mapping, yet their pipeline assumes static graphs and overlooks transition scoring or redundancy.

AutoSlim targets graph complexity at the front end. It applies supervised learning to prune low-utility transitions using features like edge score, frequency, and node centrality. Unlike static rule-based approaches, AutoSlim adapts across workloads and scales to large automata with scoring semantics. When integrated with NAPOLY+, it reduces synthesis complexity and hardware usage while preserving match correctness.

\section{AutoSlim Tool and Methodology}

AutoSlim introduces a modular and scalable software framework designed to generate, annotate, and intelligently prune automata graphs for acceleration on FPGA platforms. The toolchain supports the entire optimization workflow, from dataset generation to hardware-ready graph deployment. This section describes the AutoSlim methodology in two main stages: symbolic graph dataset generation and learning-based pruning.

\subsection{Symbolic Graph Dataset Generation}

To support machine learning-based graph pruning, we first developed a symbolic dataset generator compatible with the NAPOLY+ automata overlay format~\cite{karakchi2023napoly}. Our tool builds on the ANMLZoo format—widely used in automata processing research—but extends it by embedding scoring metrics into each transition. Each graph is composed of state transition elements (STEs), and every transition is annotated with a numerical score representing its importance, frequency, or computational cost.

The generator supports user-defined graph sizes and transition densities, enabling the creation of both synthetic and biologically inspired graph topologies. Internally, the graphs are encoded in XML format for NAPOLY+ compatibility. These XML files are then parsed and converted to CSV representations containing structured fields such as node IDs, transition degrees, and cumulative edge scores. This transformation prepares the data for feature extraction and subsequent use in training machine learning models.

\begin{algorithm}[t]
\caption{ML-based Transition Pruning Pipeline}
\footnotesize
\begin{algorithmic}[1]
\State \textbf{Input:} Folder of XML/CSV files with transitions, threshold $\theta$
\State \textbf{Output:} Pruned transition files and evaluation reports

\Procedure{PruneTransitions}{inputFolder, outputFolder, $\theta$}
    \ForAll{XML files in \texttt{inputFolder}}
        \State Parse each transition $(from, to, score)$
        \State Update node degree and total node score
        \State Export data to CSV format
    \EndFor
    \State Create labeled dataset $D = \{(x_i, y_i)\}$, where $y_i = \mathbb{1}[x_i > \theta]$
    \State Train Random Forest classifier on $D$
    \ForAll{CSV files}
        \State Classify each transition $(x_i)$ using the trained model
        \State Keep transitions with $\hat{y}_i = 1$
        \State Save pruned transitions to output CSV
        \State Generate report with state/transition count and accuracy
    \EndFor
    \State \Return Pruned files and evaluation reports
\EndProcedure
\end{algorithmic}
\end{algorithm}

\subsection{Transition Pruning via Supervised Learning}

The core of AutoSlim lies in its learning-based pruning pipeline, which reduces graph complexity by identifying and removing redundant or low-impact transitions. Using the CSV files generated from the symbolic graphs, AutoSlim constructs feature vectors for each transition. The primary feature used in the current implementation is the edge score, although additional graph features such as node degree or path centrality can be incorporated in future versions.

We employ a supervised Random Forest classifier~\cite{breiman2001random} to determine which transitions can be safely pruned without significantly affecting the graph’s matching accuracy. Labels for training are automatically derived from score thresholds, allowing flexible and dataset-specific cutoff tuning. The model is trained on a subset of transitions and evaluated on the remainder using standard cross-validation techniques. Each graph is then passed through the trained model to filter out transitions predicted to be non-essential. The pruned graphs are saved in the same format as the originals, facilitating direct comparison and deployment.

\subsection{Reporting and Evaluation Support}

AutoSlim generates a detailed report for each optimized graph that includes:
\begin{itemize}
    \item Total number of nodes and transitions before and after pruning.
    \item Transition pruning ratio and model prediction accuracy.
    \item Accept state preservation and edge coverage metrics (planned for future work).
\end{itemize}

This reporting functionality ensures traceability and repeatability, critical for evaluating pruning effectiveness across different datasets and hardware configurations.

\subsection{Hardware Deployment Integration}
In prior work, NAPOLY+ was implemented as a Verilog-based overlay architecture for automata acceleration~\cite{karakchi2023napoly, karbowniczak2024scored}. In this paper, we re-implemented the design using Vivado HLS to facilitate integration with AutoSlim and enable precise performance analysis, including latency, throughput, and resource utilization. The HLS version replicates the original datapath and control behavior, while introducing parameterized components that support varying graph sizes and pruning configurations. This design enables rapid prototyping and supports cycle-accurate simulation for evaluating AutoSlim-optimized automata graphs.
\begin{figure}[t]
    \centering
\includegraphics[width=\linewidth]{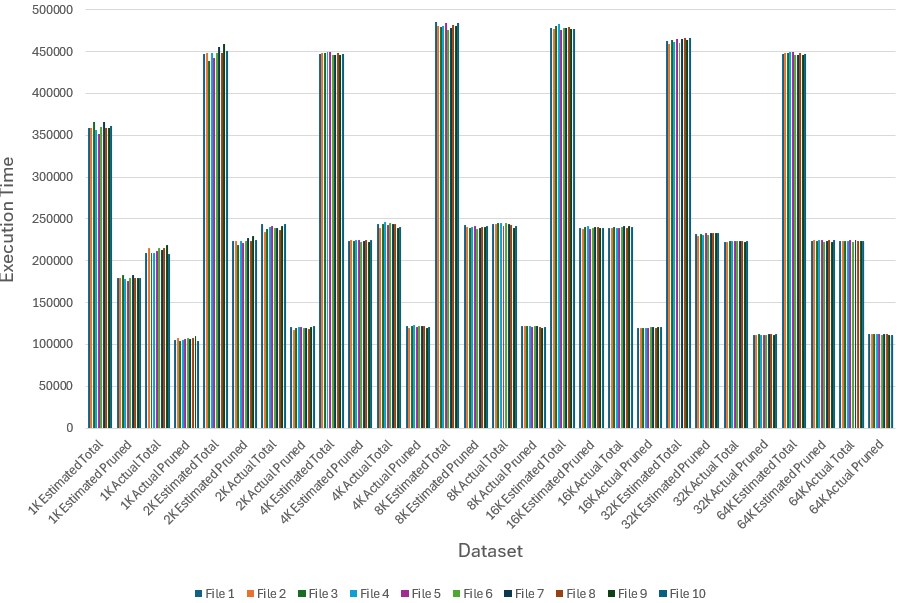}
    \caption{Execution time across ten files for each dataset size (estimated vs. actual).}
    \label{fig:execution_time}
\end{figure}

Because NAPOLY+ scales resource utilization with the number of states and transitions, the pruned graphs result in tangible reductions in hardware overhead. Our tool outputs a hardware-compatible representation of the pruned graph, which is then compiled into configuration vectors used by the HLS-based implementation of NAPOLY+. This end-to-end workflow, from symbolic graph generation to hardware deployment and performance evaluation, highlights AutoSlim’s versatility and practical value in real-world FPGA-based pattern matching applications.

\begin{figure}[t]
    \centering
\includegraphics[width=\linewidth]{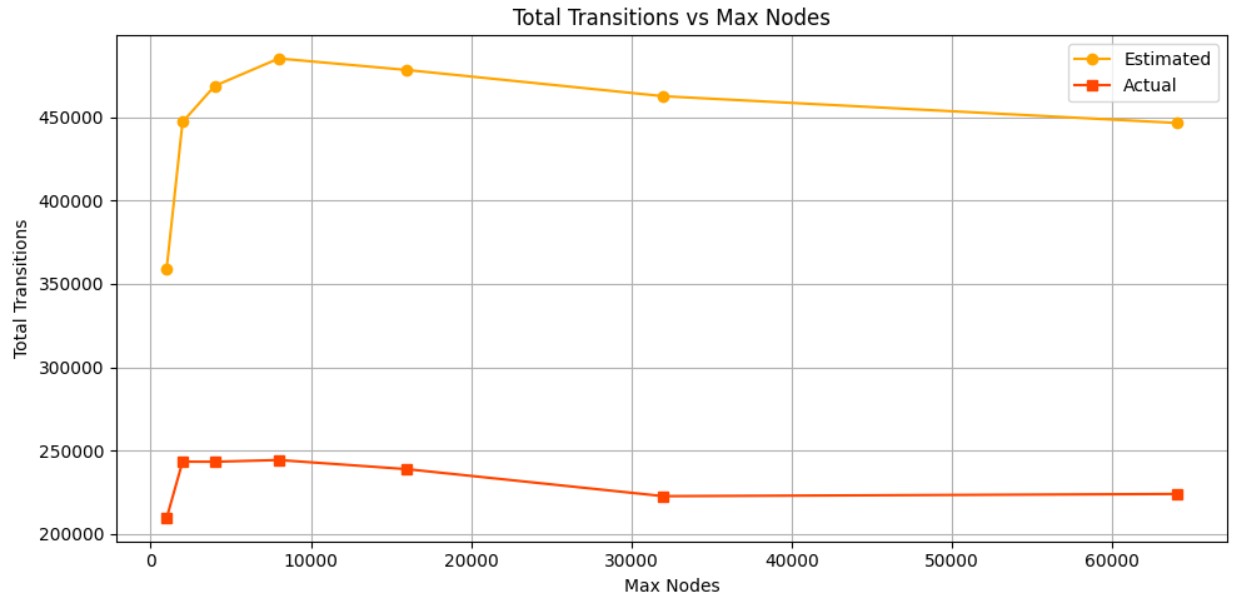}
    \caption{Total transitions vs. maximum number of nodes (estimated vs. actual).}
    \label{fig:total_transitions}
\end{figure}

\section{Evaluation}
To evaluate the performance of \textbf{AutoSlim} pruning tool, we conducted extensive experiments using synthetic datasets with increasing node counts ranging from 1K to 64K. The evaluation considers three main metrics: \emph{execution time}, \emph{total transitions}, and \emph{average transitions per node}, each measured before and after applying AutoSlim’s pruning strategy. 

\subsection{Execution Time Across Datasets}

As shown in Figure~\ref{fig:execution_time}, we evaluated the pruning execution time for ten representative files in all sizes of the data set. Each group of bars contains the execution times for the ten runs per configuration, separated into \emph{Estimated} and \emph{Actual} results. The estimated version reflects the cost predicted by AutoSlim based on heuristic scoring prior to ML-based refinement, whereas the actual version measures the final time taken after pruning. The results show that execution time generally decreases with increasing dataset size, primarily due to the sparsity induced by pruning and the reduced number of active transitions. For smaller datasets (1K to 4K nodes), execution time is higher due to denser graphs and less opportunity for pruning, while for larger datasets (16K--64K) the pruning percentage is higher. which leads to reduction in execution time. 
\subsection{Total Transitions vs. Max Nodes}

Figure~\ref{fig:total_transitions} presents the total number of transitions before and after pruning, plotted against the maximum number of nodes in the graph. The \emph{Estimated} line indicates the number of transitions derived from our heuristic graph generator, while the \emph{Actual} line shows the number of transitions retained after ML-based pruning. We observe that estimated transitions increase with graph size up to 8K nodes and then plateau, while actual transitions remain nearly flat across all sizes. This result demonstrates that AutoSlim consistently prunes irrelevant transitions regardless of scale, focusing only on semantically meaningful paths as identified by the learning model.

\subsection{Average Transitions Per Node}

\begin{figure}[t]
    \centering \includegraphics[width=0.47\textwidth]{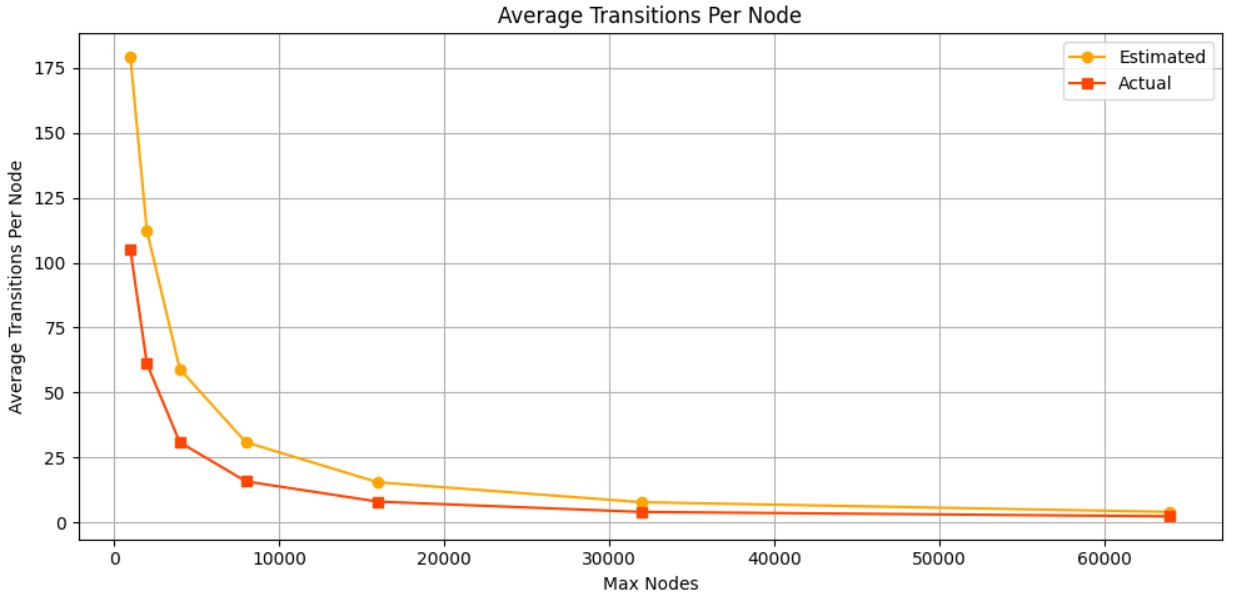}
    \caption{Average Transitions per Node Before and After Pruning}
    \label{fig:transitions_avg}
\end{figure}

To further analyze graph sparsity, Figure~\ref{fig:transitions_avg} reports the average number of transitions per node before and after pruning. As expected, the unpruned graphs are highly dense for smaller graphs (e.g., over 175 transitions per node at 1K), and become sparser as graph size increases.
Post-pruning, the average transitions per node are significantly reduced across all scales. At 1K nodes, pruning reduces the value from $\sim$180 to just above 100, and for 64K nodes, it drops below 5. This shows that AutoSlim dramatically simplifies the graph structure while preserving classification utility, enabling compact and efficient deployment on hardware platforms.

\subsection{Hardware Evaluation on Zynq UltraScale+ ZCU104}

We deployed the resulting automata graphs on the Xilinx Zynq UltraScale+ ZCU104 platform. Our experiments focused on evaluating execution latency, resource utilization (LUTs, registers, URAM), and the effect of pruning across different dataset sizes (from 4K to 64K nodes).

\subsubsection{Pruning Effectiveness}

Figure \ref{fig:execution_time} and Figure \ref{fig:transitions_avg} illustrate the reduction in execution time and average transitions per node across dataset sizes. As dataset size increases, the unpruned automata graphs exhibit a steep growth in transitions and interconnect complexity, leading to elevated execution latency and memory usage. AutoSlim's pruning consistently reduces both the number of transitions and average fanout per node, while maintaining semantic equivalence. The actual number of transitions after pruning closely aligns with machine learning predictions, confirming the effectiveness of AutoSlim's classifier in identifying low-impact edges. As shown in Figure~\ref{fig:total_transitions}, pruning achieves over 30–40\% reduction in total transitions across datasets.

\begin{figure}[h]
    \centering
\includegraphics[width=0.47\textwidth]{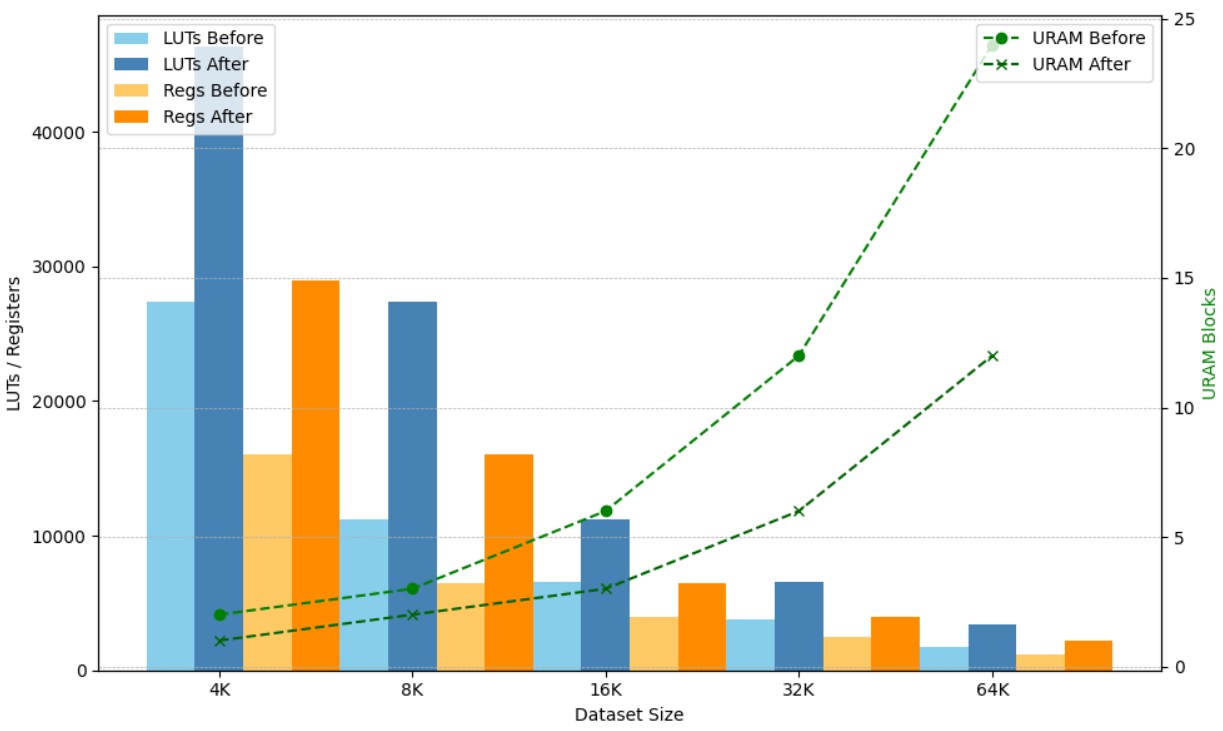}
    \caption{Resource Utilization Comparison Before and After Pruning}
    \label{fig:resource_reduction}
\end{figure}

\subsubsection{Resource Utilization and Scalability}

The pruned graphs also led to significant hardware savings. Figure~\ref{fig:resource_reduction} shows LUTs, registers, and URAM consumption before and after pruning. For instance, the 64K dataset required over 47,000 LUTs and 30K registers before pruning, which dropped to less than 7K LUTs and 2K registers after pruning. URAM usage also dropped by 50\% for all sizes.

\subsubsection{Fanout Impact Analysis}

\begin{figure}[h]
    \centering
\includegraphics[width=0.45\textwidth]{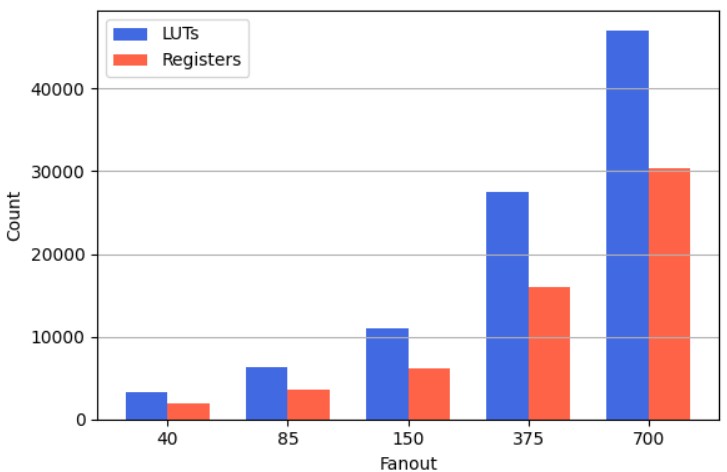}
    \caption{Resource Usage vs Fanout}
    \label{fig:resource}
\end{figure}

\begin{figure}[h]
    \centering
\includegraphics[width=0.47\textwidth]{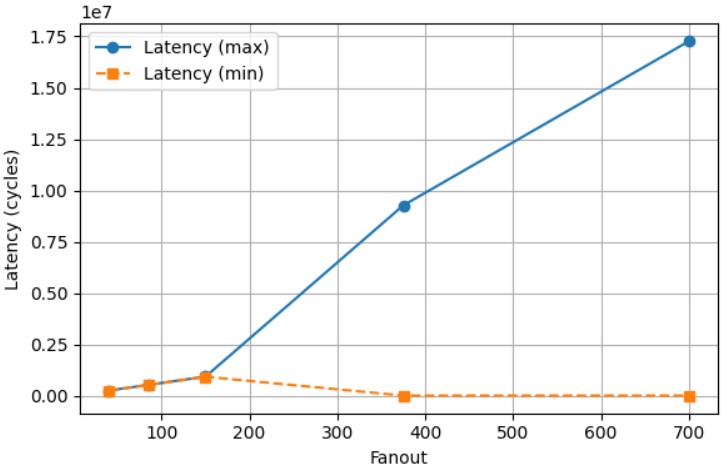}
    \caption{Latency vs Fanout (8K Dataset)}
    \label{fig:latency}
\end{figure}

To understand why smaller datasets (e.g., 4K, 8K) consumed disproportionately high resources, we conducted experiments varying the fanout on the 8K dataset. Figure \ref{fig:resource} presents the synthesis results. With higher fanout (e.g., 375 or 700), we observed a dramatic increase in logic usage and routing complexity. Although latency was reduced at high fanout due to increased parallelism, the resource demand grew significantly, particularly in LUTs and registers as shown in Figure \ref{fig:latency}.

Figure~\ref{fig:latency} shows how maximum and minimum latency values scale with fanout. While larger fanout improves parallelism (lower min latency), it also increases routing complexity and worst-case latency. AutoSlim enables a balance by reducing average fanout while preserving critical scoring paths, as observed in our experiments on the Zynq UltraScale+ platform.

\section{Conclusion}
This paper presents \textbf{AutoSlim}, a machine learning-guided framework for optimizing symbolic automata graphs deployed on FPGA accelerators. Targeting score-based Non-deterministic Finite Automata implemented using NAPOLY+, AutoSlim performs data-aware pruning to eliminate redundant transitions and unreachable nodes. By integrating structural heuristics with supervised learning (Random Forest), AutoSlim significantly reduces hardware resource usage without compromising semantic accuracy.




In future work, we plan to expand AutoSlim’s classifier to include graph-level structural features such as betweenness centrality, path entropy, and node recurrence. We also aim to explore deeper models (e.g., Deep Forests, Graph Neural Networks) and evaluate across additional hardware platforms including CGRAs and memory-centric dataflow architectures. Finally, integrating AutoSlim with automated verification pipelines and FPGA place-and-route feedback loops may enable end-to-end optimization from algorithm to silicon.

\section{Acknowledgment}
This work was supported by Office of Undergraduate Research and McNair Junior Fellowship at University of South Carolina. The authors used OpenAI’s ChatGPT to assist with language and grammar refinement. All technical content and analysis were solely developed by the authors.

\bibliographystyle{plain}
\bibliography{main}

\end{document}